\documentclass[12pt, letterpaper]{article}
\usepackage[margin=1in]{geometry}
\usepackage[symbol]{footmisc}

\usepackage[square,numbers]{natbib}
\usepackage[utf8]{inputenc} 
\usepackage[T1]{fontenc}    
\usepackage[bookmarks=false,colorlinks=true,linkcolor=blue,urlcolor=purple,citecolor=red,pdfstartview=FitH,pagebackref]{hyperref}       
\usepackage{url}            
\usepackage{booktabs}       
\usepackage{amsfonts}       
\usepackage{nicefrac}       
\usepackage{microtype}      
\usepackage{xcolor}         

\usepackage{graphicx,subfig}
\usepackage{tikz}
\usepackage{makecell}
\usepackage[linesnumbered,lined,ruled,commentsnumbered]{algorithm2e}
\let\oldnl\nl
\newcommand{\nonl}{\renewcommand{\nl}{\let\nl\oldnl}}

\usepackage{multicol,lipsum}
\usepackage{amssymb}

\usepackage{amsmath}

\usepackage{floatrow} 
\usepackage{multirow}

\usepackage{mathabx}

\newcounter{assumption}
\renewcommand{\theassumption}{A\arabic{assumption}}

\newcommand{\beq}{\begin{equation}}
\newcommand{\eeq}{\end{equation}}
\newcommand{\beqa}{\begin{eqnarray}}
\newcommand{\eeqa}{\end{eqnarray}}
\newcommand{\beqan}{\begin{eqnarray*}}
\newcommand{\eeqan}{\end{eqnarray*}}
\newcommand{\ben}{\begin{eqnarray*}}
\newcommand{\een}{\end{eqnarray*}}

\newcommand{\abs}[1]{\left\vert#1\right\vert}

\newcommand{\eps}{\varepsilon}

\newcommand{\argmax}{\mathop{\textrm{argmax}}}

\newcommand{\Ours}{{\text{SAGE}}}

\newif\ifconsiderlater
\considerlatertrue

\newif\ifSupp
\Supptrue

\ifconsiderlater
    \newcommand{\AFComment}[1]{{\color{red}{[AF: #1]}}}
    \newcommand{\KDComment}[1]{{\color{blue}{[KD: #1]}}}
    \newcommand{\NDComment}[1]{{\color{orange}{[ND: #1]}}}
    \newcommand{\LPComment}[1]{{\color{magenta}{[LP: #1]}}}
    \newcommand{\RZComment}[1]{{\color{green}{[RZ: #1]}}}
    \newcommand{\AMComment}[1]{{\color{purple}{[AM: #1]}}}
    
\else
    \newcommand{\AFComment}[1]{}
    \newcommand{\KDComment}[1]{}
    \newcommand{\NDComment}[1]{}
    \newcommand{\LPComment}[1]{}
    \newcommand{\RZComment}[1]{}
    \newcommand{\AMComment}[1]{}
\fi

\newif\ifUseBMVC
\UseBMVCfalse

\ifUseBMVC
    \newcommand\FigHeight{20}
    \newcommand\FigOffsetHeight{19}
    \newcommand\TabResultRowSpace{1.1}
    \newcommand\FigSuppHeight{13}
\else
    \newcommand\FigHeight{25}
    \newcommand\FigOffsetHeight{24}
    \newcommand\TabResultRowSpace{1.3}
    \newcommand\FigSuppHeight{15}
\fi

\newif\ifIncludeImageNet
\IncludeImageNetfalse

\title{SAGE: Saliency-Guided Mixup with \\ Optimal Rearrangements}

\author{Avery Ma$^{1,2\thanks{Work done during an internship at Samsung AI Centre Toronto}}$ \and Nikita Dvornik$^3$ \and Ran Zhang$^3$ \and Leila Pishdad$^4\thanks{Work done while at Samsung AI Centre Toronto}$ \and Konstantinos G. Derpanis$^{2,3,5}$ \and Afsaneh Fazly$^3$}
\date{%
    $^1$University of Toronto\quad$^2$Vector Institute\\[1ex]
    $^3$Samsung AI Centre Toronto\quad$^4$Borealis AI\quad$^5$York University
}

\begin{document}

\maketitle

\vspace{-10pt}
\begin{abstract}
Data augmentation is a key element for training accurate models by reducing overfitting and improving generalization.
For image classification, the most popular data augmentation techniques range from simple photometric and geometrical transformations, to more complex methods that use visual saliency
to craft new training examples.
As augmentation methods get more complex, their ability to increase the test accuracy improves, yet, such methods become cumbersome, inefficient and lead to poor out-of-domain generalization, as we show in this paper.
This motivates a new augmentation technique that allows for high accuracy gains while being simple, efficient (i.e., minimal computation overhead) and generalizable.
To this end, we introduce \textbf{Sa}liency-\textbf{G}uided Mixup with Optimal R\textbf{e}arrangements (SAGE), which creates new training examples by rearranging and mixing image pairs using visual saliency as guidance.
By explicitly leveraging saliency, \Ours{} promotes discriminative foreground objects and produces informative new images useful for training.
We demonstrate on CIFAR-10 and CIFAR-100 that \Ours{} achieves better or comparable performance to the state of the art while being more efficient.
Additionally, evaluations in the out-of-distribution setting, and few-shot learning on mini-ImageNet, show that \Ours{} achieves improved generalization performance without trading off robustness.
%
%
Our source code is available at \url{https://github.com/SamsungLabs/SAGE}.

\end{abstract}
\section{Introduction}
\label{sec:intro}

\begin{figure}
\captionsetup[subfigure]{labelformat=empty}
\centering
\subfloat[Batch]{\includegraphics[height=\FigHeight mm]{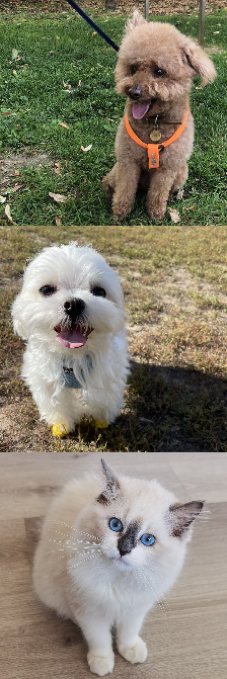}}
\hfill
\hfill
\subfloat[Mixup\cite{zhang2018mixup}]{\includegraphics[height=\FigHeight mm]{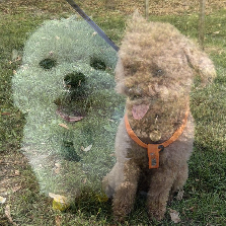}}
\hfill
\subfloat[CutMix\cite{yun2019cutmix}]{\includegraphics[height=\FigHeight mm]{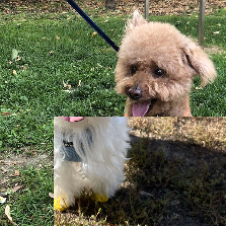}}
\hfill
\subfloat[SaliencyMix\cite{uddin2020saliencymix}]{\includegraphics[height=\FigHeight mm, trim=10px 10px 10px 10px, clip]{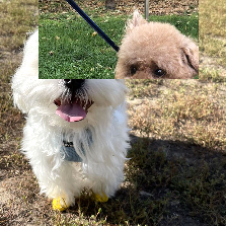}}
\hfill
\subfloat[Puzzle Mix\cite{kim2020puzzle}]{\includegraphics[height=\FigHeight mm]{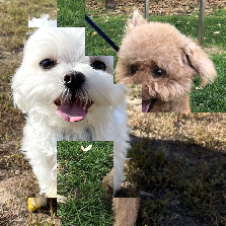}}
\hfill
\subfloat[Co-Mixup\cite{kim2020co}]{\includegraphics[height=\FigHeight mm]{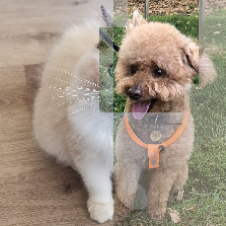}}
\hfill
\subfloat[\Ours{} (ours)]{\includegraphics[height=\FigHeight mm]{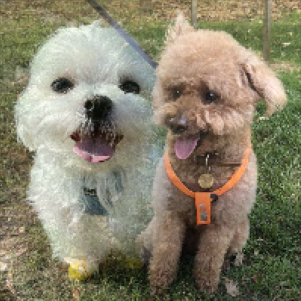}}
\caption{\textbf{Comparison of data augmentation methods.} Thanks to the saliency-guided mixing and image rearrangements, \Ours{} produces more meaningful and informative scenes, as verified in our experiments. 
}
\label{fig:motivation}
\end{figure}

Data augmentation (DA) methods synthetically expand a dataset by applying transformations on the available examples, with the goal of
reducing overfitting and improving generalization in models trained on these datasets.
In computer vision, conventional DA techniques are typically based on random geometric (translation, rotation and flipping) and photometric (contrast, brightness and sharpness) transformations~\citep{simonyan2015very, lecun1998gradient, cubuk2019autoaugment, cubuk2020randaugment}.
While these techniques are already effective, they merely create slightly altered copies of the original images and thus introduce limited diversity in the augmented dataset.
A more advanced DA~\cite{zhang2018mixup,yun2019cutmix} combines multiple training examples into a new image-label pair.
By augmenting both the image and the label space simultaneously, such  approaches greatly increase the diversity of the augmented set.
Consequently, they substantially improve model generalization, without any efficiency overhead, due to their simplicity.
Nonetheless, these DA approaches are agnostic to image semantics
; they ignore object location cues, and as a result may produce ambiguous scenes with occluded distinctive regions (see Figure~\ref{fig:motivation}, Mixup  \cite{zhang2018mixup} and CutMix  \cite{yun2019cutmix}).

To account for such shortcomings, a new line of work~\cite{kim2020puzzle,kim2020co,gong2021keepaugment,uddin2020saliencymix} proposes to explicitly use visual saliency~\cite{simonyan2013deep} for data augmentation.
Typically, a saliency map contains the information about the importance of different image regions for the downstream task.
As a result, saliency maps implicitly contain information about objects, their locations and, crucially, about the ``informativeness'' of image regions.  Previous methods~\cite{kim2020puzzle,kim2020co,uddin2020saliencymix} take full advantage of the saliency information, and formulate data augmentation as a saliency maximization problem.
Given training image patches,
their augmentation ``assembles'' a new image of high visual saliency.
This approach greatly improves the test accuracy; however, this comes with a large computation overhead due to the need to maximize saliency at every training step.
Moreover, as the augmented images are composed of patches, the resulting scenes are often unrealistic (see Puzzle Mix, Co-Mixup and SaliencyMix in Figure~\ref{fig:motivation}), which leads to poor out-of-distribution generalization, as shown later in our experiments.
In summary, the existing data augmentation techniques can \emph{either} i) boost the test accuracy, \emph{or} ii) produce a robust model with little computational overhead; there are no methods that can do both.

To address the aforementioned drawbacks, we propose a new augmentation -- \textbf{Sa}liency-\textbf{G}uided Mixup with Optimal R\textbf{e}arrangements (SAGE) -- that provides both high accuracy and robustness, and has minimal computation overhead. 
\Ours{} is a simple and effective DA technique that uses visual saliency to perform optimal image blending
at each spatial location, and optimizes the relative image position such that the resulting visual saliency is maximized.
Given two images and their saliency maps, \Ours{} mixes the images together, such that at each spatial location, the contribution of different images to the mix is proportional to their saliency in that location.
The corresponding label is also obtained by interpolating the original labels based on the saliency of the corresponding images.
To maximize the resulting saliency of the mix, we find an optimal relative arrangement of the two images prior to the mixing stage.
As a result, \Ours{} produces smooth and realistic images with clear and distinct foreground objects (see Figure~\ref{fig:motivation}), unlike other augmentation techniques.
Thanks to our efficient implementation, \Ours{} has virtually no computation overhead beyond obtaining the saliency information.
Furthermore, our computations are partially shared between the saliency masks and the training gradients, which further decreases the amortized training time.

\noindent{\bf Contributions. } We make the following three contributions:
(i) We introduce \Ours{}, a DA method to generate novel training examples by mixing image pairs based on their visual saliency, which promotes discriminative foreground objects in the mix. (ii)
\Ours{} achieves test accuracy better than or comparable to state-of-the-art augmentation techniques, without incurring significant computation overhead.
(iii) Through robustness evaluations on perturbed test data, we show that \Ours{} improves test accuracy without trading off robustness. 
\section{Related Work}
\label{sec:related}
In this section, we review data augmentation techniques that go beyond simple geometrical and color transformations to improve generalization.
A popular approach is to synthesize new training input-output pairs by combining information from multiple raw samples.
Mixup~\cite{zhang2018mixup} creates a new image-label pair by linearly interpolating both the input and output space.
In contrast, Manifold Mixup~\cite{verma2019manifold} and HypMix~\cite{sawhney2021hypmix} apply interpolation at the feature level.
Others create new training samples by ``copy-pasting'' patches from one image to another~\cite{yun2019cutmix, ghiasi2021simple, fang2019instaboost}.
This class of methods is very efficient and simple to implement.
However, a common drawback of these approaches is that they do not take image semantics into account when performing augmentation.
This potentially encourages the model to generalize using completely irrelevant information from the new training data, leading to inferior generalization.

To address this problem, recent work explicitly uses visual saliency information in the DA process.
KeepAugment~\cite{gong2021keepaugment} leverages input saliency to improve existing DA techniques, e.g., Cutout~\cite{devries2017improved}, by always keeping the important regions untouched during augmentation.
SaliencyMix~\citep{uddin2020saliencymix} improves CutMix~\citep{yun2019cutmix} by selecting a patch around the peak salient pixel location in the source image and mixing it with the target image.
Puzzle Mix formulates DA as an optimization problem, where the objective balances saliency maximization, local smoothness and 
the optimal transport between data pairs~\cite{kim2020puzzle}. 
Co-Mixup~\cite{kim2020co} extends this idea by encouraging the diversity of the augmentation when mixing a collection of inputs, and thus further complicates the optimization objective.
The need to solve the optimization problem at every step significantly slows down the training, which may be prohibitive in some situations. Our saliency-guided method not only reduces this computational overhead, but also generates more plausible augmented images that result in improved test accuracy and out-of-distribution generalization. 
\vspace{-3pt}
\section{Technical Approach}
\label{sec:method}

\begin{figure}[t]
\centering
\includegraphics[trim=20 120 110 45,clip,width=0.99\textwidth]{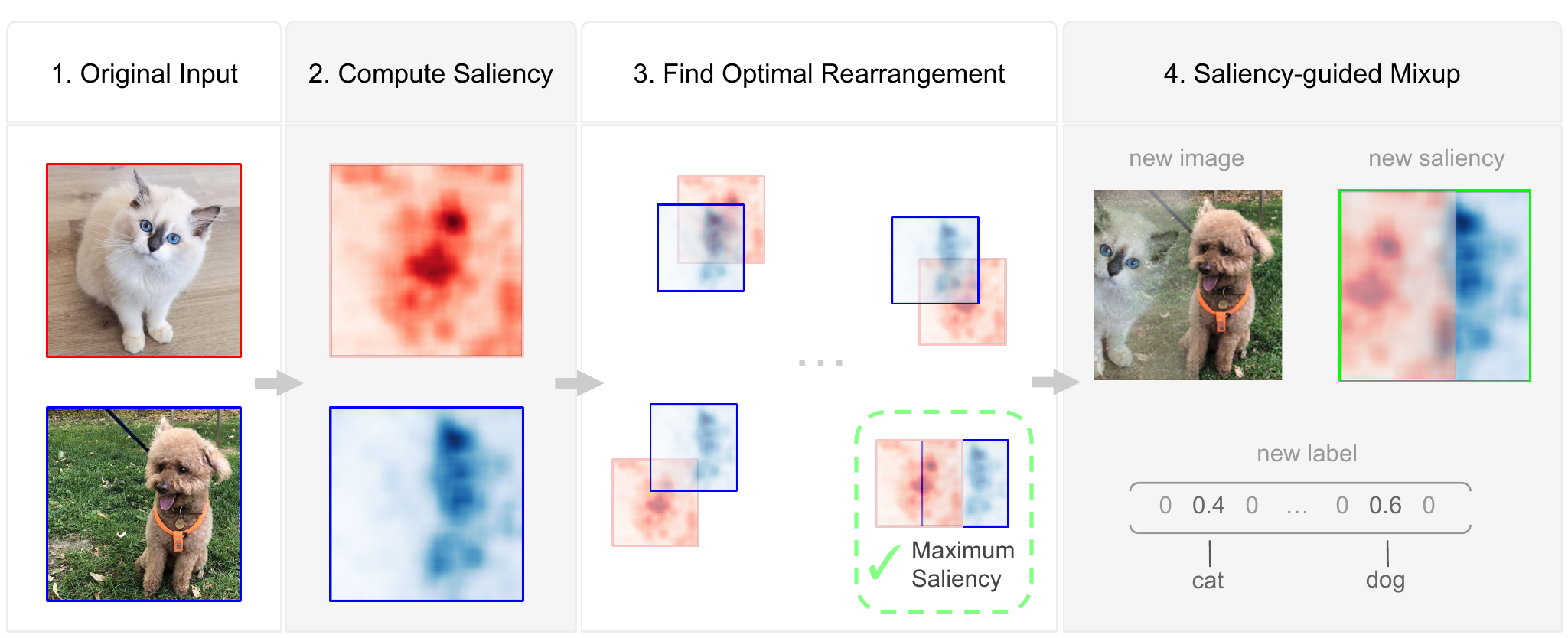}
\caption{\textbf{\Ours{} overview}.
Given the original images, we first compute saliency maps. Next, we find the best rearrangement of the images that maximizes the total saliency (in the green box).  
Finally, we use our saliency-guided Mixup to fuse the overlapping image parts and derive the new label.
As a result, \Ours{} produces smooth, realistic and informative scenes.
\label{fig:pipeline}
}
\vspace{-8pt}
\end{figure}

The main idea behind \Ours{} is to synthesize novel images (with their labels) by blending pairs of training samples, using spatial saliency information as guidance for optimal blending.
As illustrated in Fig.~\ref{fig:pipeline}, our method consists of three independent components: i) saliency mask generation (Sec.~\ref{sec:saliency}), ii) the ``Optimal Rearrangement'' module (Sec.~\ref{sec:mix_with_shift}), and iii) the ``Saliency-guided Mixup'' module (Sec~\ref{sec:mixup}).
All chained together, they form our \Ours{} approach.
Below, we elaborate on each of the components and conclude with a discussion on the efficiency of our pipeline in Sec.~\ref{sec:efficiency}.

\subsection{Computing Saliency Maps}\label{sec:saliency}
We define the saliency of each image pixel as its importance in making the correct prediction, using a given vision model. 
More formally, we are given a training sample, $(x, y)$, where $x \in \mathbb{R}^{d \times d \times 3}$ is an RGB image and $y \in \mathbb{R}^C$ is the corresponding one-hot label vector, a
classifier, $f_{\theta}(\cdot)$, that is the current partially trained model, and our task loss, $\ell(f_{\theta}(x), y)$, measuring the discrepancy between the classifier's output and the true label.
We define the saliency, $s \in \mathbb{R}^{d \times d}$, as the magnitude of the gradient with respect to the input image,
\begin{equation}\label{eq:saliency}
    s(x) = \abs{ \nabla_{x} \ell(f_{\theta}(x), y) }_{l_{2,3}},
\end{equation}
where $\abs{\cdot}_{l_{2,3}}$ denotes the $l2$-norm along the third (color) dimension.
In practice, the saliency map tends to focus on the foreground objects useful for classification and ignores irrelevant background.
Note that our saliency definition differs from others~\cite{simonyan2013deep,selvaraju2017grad} in that we consider the gradient of the full loss, while previous work consider the gradient of the ground-truth class activation with respect to the input image.
We find that our definition is advantageous for data augmentation, and additionally allows for more efficient training, as detailed in Sec.~\ref{sec:efficiency}.

\subsection{Saliency-guided Mixup}\label{sec:mixup}
\begin{figure}[t]
\centering
\includegraphics[trim=25 100 10 29,clip,width=0.9\textwidth]{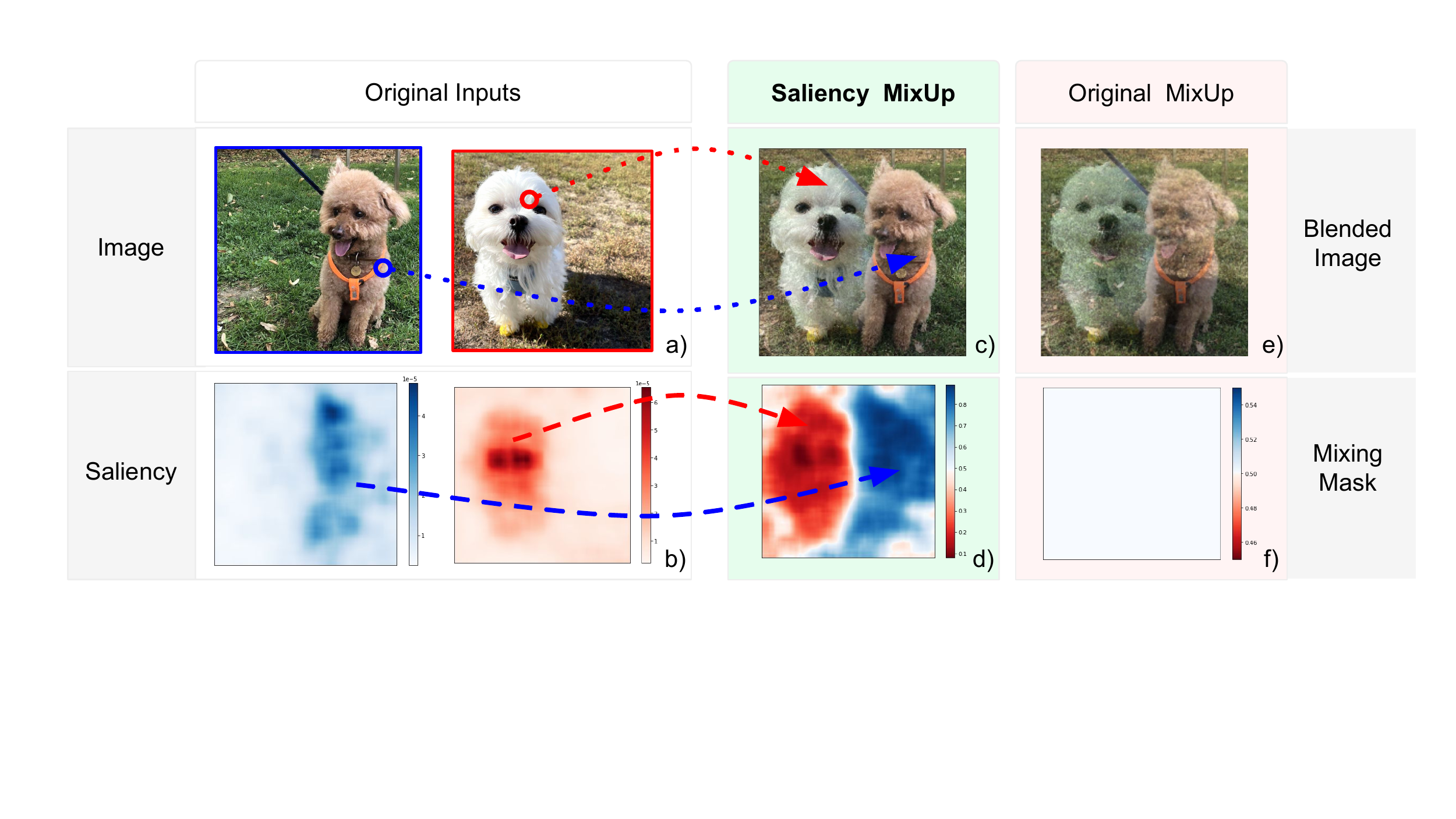}
\caption{\textbf{Comparison between Saliency-guided Mixup and original Mixup}.
Given the a) original images with b) saliency maps, our Saliency Mixup computes d) the Mixing Mask $M$ (given by Eq.~\ref{eq:SGM}) based on the relative saliency of the inputs. The values of $M$ are represented with a heatmap; blue areas indicate stronger contribution of image 1, red areas correspond to image 2 being more prominent and pale areas indicate more uniform blending. 
Consequently, salient regions from different images contribute to different locations and result in a realistic, informative output c).
In contrast, the original Mixup produces f) a uniform mixing mask (at $\lambda = 0.5$), which results in e) an unrealistic and unclear image.}
\label{fig:saliency-mixup}
\vspace{-8pt}
\end{figure}

Before describing our Saliency-guided Mixup, we revisit the original Mixup~\cite{zhang2018mixup}.
Mixup 
creates a new training sample, $x'$, by linearly mixing pairs of training samples, $x_1$ and $x_2$, i.e., $x' = \lambda \cdot x_0 + (1-\lambda)\cdot x_1$, and their corresponding labels,  i.e., $y' = \lambda \cdot y_0 + (1-\lambda)\cdot y_1$, 
where $\lambda \in [0, 1]$.
While simple and effective, Mixup has a notable drawback, namely it 
ignores 
the image semantics.
That is, at every pixel location, the contribution of $x_1$ and $x_2$ to the final image is constant. 
As Fig.~\ref{fig:saliency-mixup} (e) shows, this may lead to prominent image regions being suppressed by the background, which is not ideal for data augmentation~\cite{kim2020puzzle,kim2020co}.

To address this shortcoming, we propose Saliency-guided Mixup, where at every image location in $x'$, the mixing ratio between $x_1$ and $x_2$ is different, defined by the saliency of the corresponding image regions.
More formally, given two images, $x_1$ and $x_2$, and their saliency maps, $s_1$ and $s_2$, we craft a 2D mixing mask, $M \in \mathbb{R} ^ {d \times d}$
, and use it to mix the images: 
\begin{equation}\label{eq:SGM}
    x' = M \odot x_1 + (1 - M) \odot x_2, \ ; \  M = \frac{\tilde{s}_1}{\tilde{s}_1 + \tilde{s}_2 + \zeta},
\end{equation}
where $x' \in \mathbb{R} ^ {d \times d \times 3}$,
$\tilde{s}_1$ and $\tilde{s}_2$ are spatially-normalized and Gaussian-smoothed saliency maps, $\zeta$ is a scalar hyperparameter used to avoid division-by-zero and $\odot$ denotes element-wise product. 
That is, the elements in $M$ are defined as the saliency ratio in different images at the same location.
This means that, at any given location, more prominent regions of one image will suppress less salient regions of the other image in the final blend, $x'$.
This strategy largely resolves the issue with the original Mixup and leads to more informative augmentation (see Fig~\ref{fig:saliency-mixup} (e)).
Lastly, we mix the labels using $y' = \gamma \cdot y_0 + (1-\gamma) \cdot y_1$, where $\gamma$ is the mean of the mixing mask, $M$.

Saliency-guided Mixup, Eq.~\ref{eq:SGM}, is most suitable for mixing images that have salient regions in distinct locations.
When the maximally salient regions in both images spatially overlap, the mask, $M$, tends to suppress one or both objects, which leads to uninformative new scenes.

\subsection{Optimal Rearrangements via Saliency Maximization}\label{sec:mix_with_shift}
\begin{figure}[t]
\captionsetup[subfigure]{labelformat=simple,labelsep=period}
\centering     
\subfloat[Total saliency $v(\tau_1) = 0.48$]{\label{fig:offset_1}\includegraphics[height=\FigOffsetHeight mm]{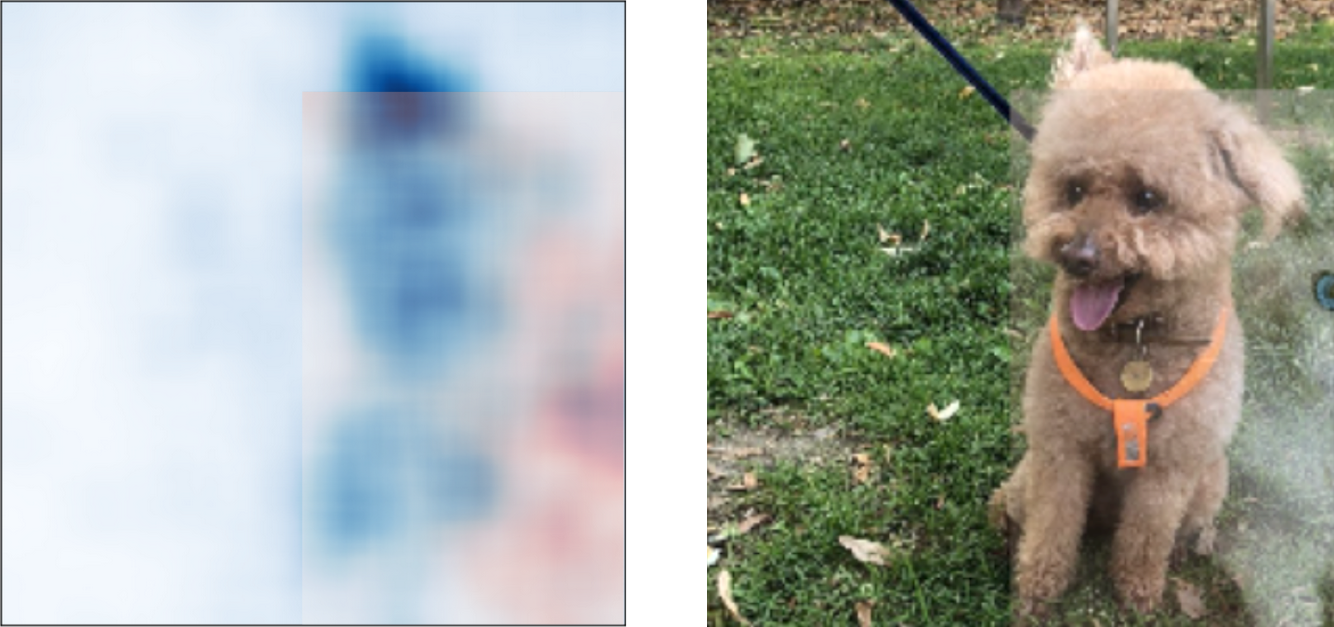}}\
\hfill
\subfloat[Total saliency $v(\tau_1) = 0.57$]{\label{fig:offset_2}\includegraphics[height=\FigOffsetHeight mm]{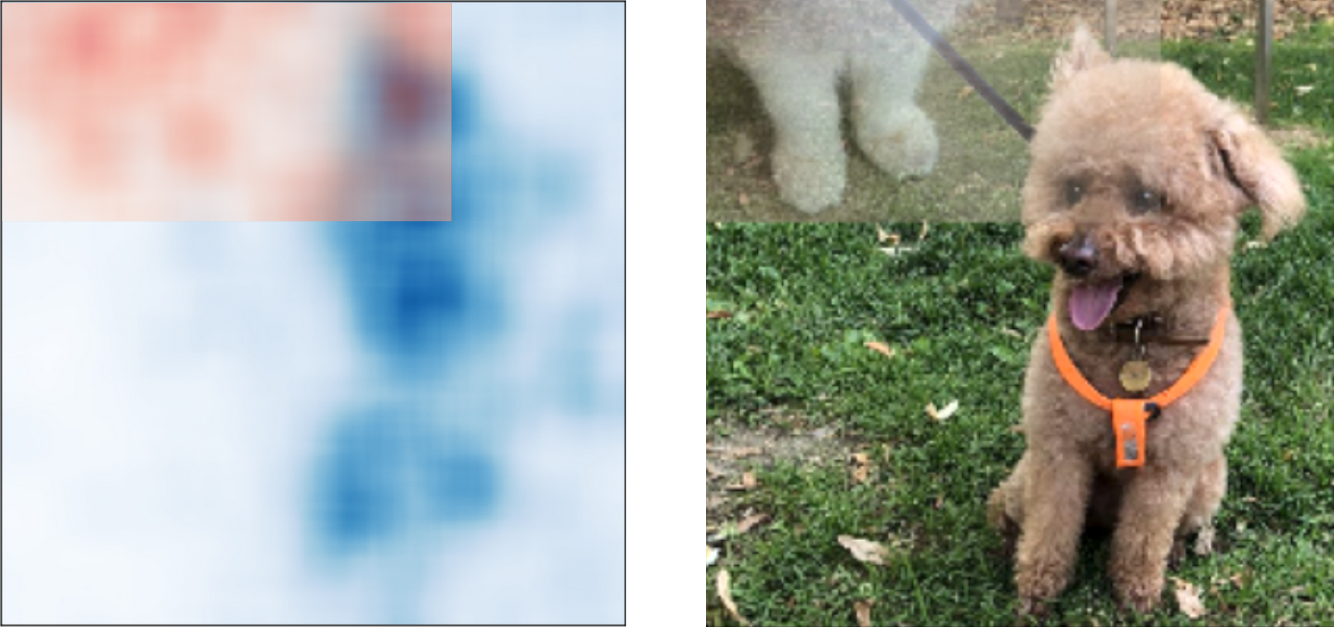}}\
\hfill
\subfloat[Max saliency $v(\tau_1) = $\textbf{ 0.72}]{\label{fig:offset_best}\includegraphics[height=\FigOffsetHeight mm]{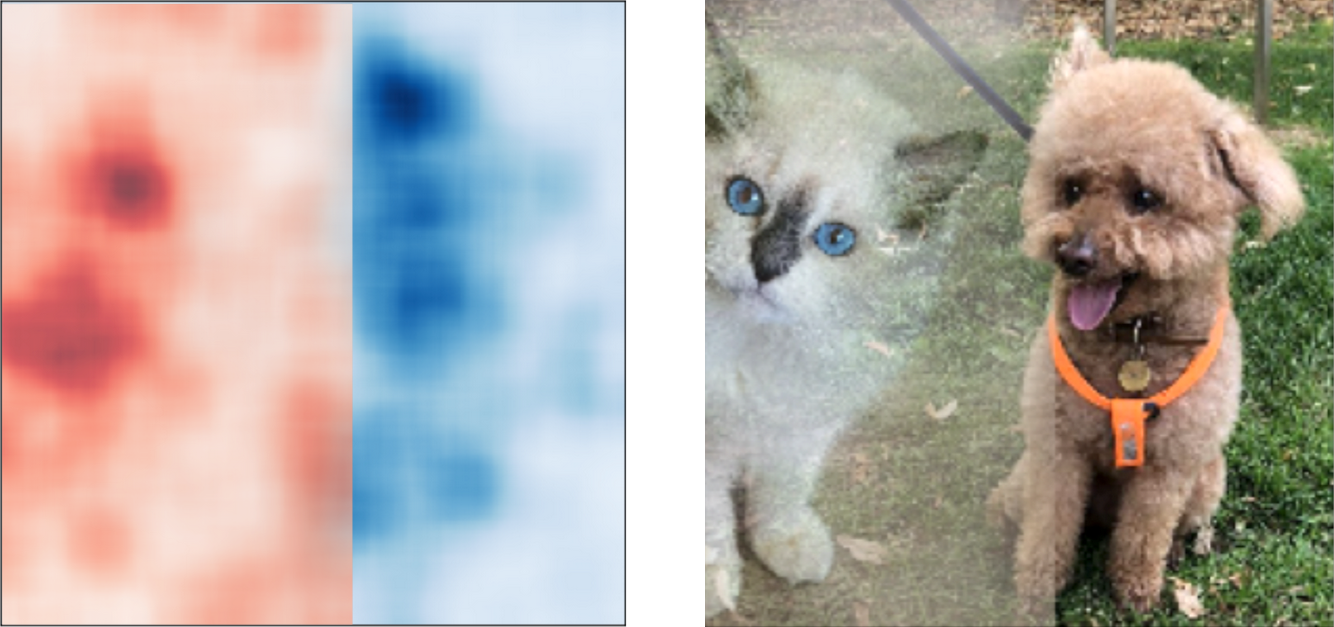}}
\caption{\textbf{Possible rearrangements}. In each example, the saliency map corresponding to the rearrangement is shown on the left, the corresponding image (after applying Saliency-guided Mixup) is on the right.
The rearrangement maximizing the total saliency is shown in c); clearly, it results in a denser mixed saliency, and produces a more informative image.
\vspace{-10pt}
}
\label{fig:offset_comparison}
\vspace{-5pt}
\end{figure}

To produce highly-informative augmentations with Eq.~\ref{eq:SGM}, even when both images have overlapping salient regions, we propose to shift one image relative to the other prior to mixing.
Our objective is to find the shift that maximizes the resulting image saliency.
An example of such rearrangements with the resulting augmentations are shown in Fig.~\ref{fig:offset_comparison}.
In the following, we formalize this shifting process and describe a solution for finding the best rearrangement.


We define the translation operator that shifts a tensor $z$ by $\tau=(\tau_i$, $\tau_j$) pixels as 
\begin{equation}\label{eq:rearrange}
    \mathcal{T}(z, \tau)[i, j] =
    \begin{cases}
         z[i - \tau_i, j - \tau_j], \ \text{if} \ i - \tau_i \in [0, d-1], j - \tau_j \in [0, d-1] &\\
         0, \ \text{otherwise}&
    \end{cases},
\end{equation}
where $z[i, j]$ is the value of $z$ at the location $(i, j)$.
Essentially, translation $\mathcal{T}$ shifts all the values in the tensor by the given offset, $\tau$, and zero-pads the empty space.

To quantify how successful a given rearrangement  is in resolving  the saliency overlap, we measure the total saliency~\cite{kim2020puzzle} after the rearrangement.
For a given rearrangement, $\tau$, the total saliency, $v(\tau) \in \mathbb{R}$, is defined as follows:
\begin{equation}\label{eq:total-saliency}
    v(\tau) = \sum_{i,j} \big [ M^{\tau} \odot \tilde{s}_1 + (1 - M^{\tau}) \odot \mathcal{T}(\tilde{s}_2, \tau) \big ], 
\end{equation}
where $\mathcal{T}(\tilde{s}_2, \tau)$ is the saliency $\tilde{s}_2$ translated by $\tau$ and $M^{\tau}$ is the mixing mask (Eq.~\ref{eq:SGM}) computed with $\tilde{s}_1$ and $\mathcal{T}(\tilde{s}_2, \tau)$.
Essentially, the scalar $v(\tau)$ captures the total saliency after the rearrangement (Eq.~\ref{eq:rearrange}) and fusion (Eq.~\ref{eq:total-saliency}) of the individual saliency tensors.
Intuitively, larger total saliency values imply smaller overlap between the salient regions in the shifted images, $x_1$ and $\mathcal{T}(x_2, \tau)$, and suggests that the resulting mix is more informative.
Thus, it is reasonable to look for a rearrangement that maximizes the total saliency.
To this end, we propose to find the optimal rearrangement (offset), $\tau^{\ast}$, by solving the following: $\tau^{\ast} = \argmax_{\tau \in \mathcal{O}} v(\tau)$, where $\mathcal{O}$ is the space of all possible offsets (shown in Fig.~\ref{fig:pipeline}, step 3).

Finally, we use the obtained optimal rearrangement to generate the augmented sample, $x'$.
This is done by applying our Saliency-guided Mixup to the rearranged image pair (shown in Fig.~\ref{fig:pipeline}, step 4), i.e., simply plugging the images $x_1$ and $\mathcal{T}(x_2)$ with the corresponding saliency $\tilde{s}_1$ and $\mathcal{T}(\tilde{s}_2)$ into Eq.~\ref{eq:SGM}.
The exact data augmentation algorithm is detailed in the supplement.

\subsection{Discussion}\label{sec:efficiency}
One of the advantages of \Ours{} over other saliency-based augmentations (e.g.,~\cite{kim2020puzzle,kim2020co}) is its efficiency.
Here, we elaborate on our pipeline design choices and discuss their complexity.
\\
\textbf{Saliency-guided Mixup.} Compared to the original Mixup blending step, our Saliency-guided Mixup (Sec~\ref{sec:mixup}) adds a simple element-wise multiplication by the mixing mask.
The cost of this operation is negligible to our model's runtime. 
\\
\textbf{Optimal Rearrangements.}
As described in Sec.~\ref{sec:mix_with_shift}, to arrive at our final mixture, we 
consider all possible rearrangements and select the one maximizing the total saliency, Eq.~\ref{eq:total-saliency}.
The number of rearrangements grows quadratically with image size and soon becomes the bottleneck.
To keep our method efficient, we randomly sample a small portion of all possible arrangements (1\% in all experiments), 
and search among them.
In our experiments, this does not affect classification performance, while greatly improving efficiency.
\\
\textbf{Saliency Computation.}
Computing saliency requires an extra forward and backward pass of the model.
When the existing works~\cite{kim2020puzzle,kim2020co} compute saliency masks, they discard all the intermediate computations and only use the mask itself for DA, which essentially doubles the training time.
In contrast, \Ours{} saves the gradients, $g_s$, with respect to the model parameters, obtained in the backward pass of saliency computations.
These gradients can be combined with the standard gradients, $g_a$, computed on \Ours{}-augmented images to perform the final model update with $g = \eta \cdot g_s + (1-\eta) \cdot g_a$, where $\eta \in [0, 1]$.
The hyperparameter, $\eta$,
effectively controls how much information from the original images is used for updates versus that of the augmented images.
This trick allows us to amortize the saliency computations, and reuse the intermediate results for the model updates.
Note that this is only possible thanks to our saliency definition (Eq.~\ref{eq:saliency}), which differs from the classical one~\cite{simonyan2013deep}.

\section{Experiments}
\label{sec:experiments}
We demonstrate the advantage of \Ours{} in image classification in Sec.~\ref{sec:results}. Sec.~\ref{sec:robustnes} evaluates \Ours{} in out-of-distribution generalization, Sec.~\ref{sec:runtime} analyzes the efficiency of our pipeline and Sec.~\ref{sec:ablation} presents an ablation study of \Ours{}'s components. 
Our implementation is largely based on the publicly available repository of Puzzle Mix\footnote{\url{https://github.com/snu-mllab/PuzzleMix}}.  

\begin{table}[t]
\begin{center}
\renewcommand{\arraystretch}{\TabResultRowSpace}
\renewcommand{\tabcolsep}{0.8mm}
\footnotesize
\begin{tabular}{ llllllllll } 
Dataset & Model         & Vanilla    & Mixup   & CutMix   & Manifold  & SaliencyMix  & Puzzle Mix          & Co-Mixup          & \Ours{} \\
\Xhline{2\arrayrulewidth}
CIFAR-10   & PreActResNet18   & 95.07      & 95.97   & 96.27    & 96.28     & 96.15     & \underline{96.62}   & 96.23             & \textbf{96.95} \\ 
CIFAR-100  & PreActResNet18   & 76.8       & 77.40.  & 78.96    & 78.51     & 78.85     & 79.65               & \underline{79.68} & \textbf{79.91} \\ 
CIFAR-100  & WRN16      & 78.55      & 79.83.  & 80.03    & 79.77     & 80.16     & \textbf{80.73}      & 80.42             & \underline{80.45} \\ 
CIFAR-100  & ResNext29  & 78.77      & 78.23.  & 77.43    & 77.97     & 78.89     & 79.20               & \underline{80.27} & \textbf{80.35} \\
\Xhline{2\arrayrulewidth}
\hline
\end{tabular}
\end{center}
\vspace{-12pt}
\caption{\textbf{Image classification accuracy.} 
CIFAR-10 and CIFAR-100 results are obtained by averaging over three independent training runs.
The best numbers are in bold and the second best numbers are underlined.
}
\label{table:classification_accuracy}
\end{table}

\vspace{5pt}
\subsection{Image Classification}
\label{sec:results}

Following previous work~\cite{kim2020co}, 
we perform evaluations on the CIFAR-10~\cite{krizhevsky2009learning} and CIFAR-100~\cite{krizhevsky2009learning} datasets with the PreActResNet18~\cite{he2016identity}, ResNext29~\cite{xie2017aggregated} and WideResNet16~\cite{zagoruyko2016wide} architectures.
%
For all datasets and models, we follow the optimization schedule described in Puzzle Mix and Co-Mix; training and model details are included in the supplement.
For a comprehensive comparison, we use the following DA baselines:
(i) Vanilla, i.e., standard data augmentation only, which includes random cropping and horizontal flips,
(ii) Mixup~\cite{zhang2018mixup},
(iii) CutMix~\cite{yun2019cutmix},
(iv) Manifold~\cite{verma2019manifold},
(v) SaliencyMix~\cite{uddin2020saliencymix},
(vi) Puzzle Mix~\cite{kim2020puzzle}
and (vii) Co-Mixup~\cite{kim2020co}.
Note that all the baseline methods are applied on top of the standard data augmentation.
Following previous work~\cite{kim2020puzzle, kim2020co}, we report the results averaged over three independent training runs.

Table~\ref{table:classification_accuracy} summarizes the comparison of \Ours{} to the baselines, pointing to two key observations.
First, the DA techniques utilizing saliency (i.e., SaliencyMix, Puzzle Mix, Co-Mixup and \Ours{}) substantially outperform other non-saliency-based variants across almost all datasets and architectures. 
This clear improvement demonstrates that using image semantics for data augmentation leads to better generalization on the test set.
Second, among saliency-based methods, \Ours{} is consistently the best on CIFAR-10; on CIFAR-100, \Ours{} outperforms Puzzle Mix and Co-Mixup on PreActResNet18 and ResNext29, and has comparable performance on WideResNet. \Ours{} also outperforms SaliencyMix on all tested architectures on both datasets.
We attribute the advantage of \Ours{} to the fact that our augmented images are smoother and more realistic, combining the advantages of Mixup and the saliency-based methods.
This is despite the fact that Puzzle Mix and Co-Mixup are explicitly optimizing for maximum saliency, and have considerably more computational overhead.

\vspace{5pt}
\subsection{Out-of-distribution Generalization and Few-shot Adaptation}\label{sec:robustnes}
It is known that different DA techniques may lead to similar test accuracy improvements but have drastically different behavior on out-of-distribution (OOD) data~\cite{verma2019manifold}.
This phenomenon is attributed to the difference in the quality of the learned representation.
Therefore, to further evaluate our approach, we consider generalization in the OOD setting.

In our evaluation, we test the OOD generalization in two scenarios: using corrupted test images (with Gaussian noise or adversarial perturbations~\cite{szegedy2014intriguing}) or evaluating generalization to new categories in a few-shot setup~\cite{vinyals2016matching}.
More specifically, we test against three different perturbations:
i) Gaussian noise with zero mean and variance of 0.01,
ii) $\ell_{\infty}$-norm bounded attack generated using the Fast Gradient Sign Method (FGSM) \cite{goodfellow2014explaining} with $\eps = \frac{8}{255}$
and iii) $\ell_2$-norm bounded attack crafted with Fast Gradient Method (FGM)  \cite{goodfellow2014explaining} with $\eps = 0.5$.
Our choice of the attacks and $\eps$ follows the standard practice used with the robustness benchmarks~\cite{croce2020robustbench}.
To evaluate few-shot adaptation capabilities of our model and test how well the learned representations transfer to novel categories, we perform few-shot classification on the mini-ImageNet dataset~\citep{vinyals2016matching}.
Additional details are provided in the supplement. 

To summarize the performance on all three OOD benchmarks, we average the accuracy across the benchmarks, and get a single score quantifying model robustness.
Figure~\ref{fig:cifar100_robustness_comparison} plots the average OOD accuracy on CIFAR-100, against the standard accuracy on the original test set.
We observe a striking difference in the robustness characteristics across different DA methods.
Notably, models trained using \Ours{} are much less sensitive to out-of-distribution shifts compared to the two other saliency-based methods, i.e., Puzzle Mix and Co-Mixup, despite comparable test accuracy improvements.
Moreover, the models trained with CutMix, Puzzle Mix and Co-Mixup have worse OOD performance compared to Vanilla training.
These methods produce augmentations with unnatural patch-like patterns, which likely leads to unwanted properties of the learned representations.
In contrast,  Mixup and \Ours{} fuse images in a homogeneous way, leading to models more robust to various input perturbations.
Please refer to the supplement for the full table of results and CIFAR-10 experiments. 

\begin{table}[t]
\begin{center}
\renewcommand{\arraystretch}{1}
\renewcommand{\tabcolsep}{2mm}
\small
\begin{tabular}{ cccccccccc } 
Vanilla & Mixup  & CutMix &  SaliencyMix  &  Puzzle Mix  & Co-Mixup & \Ours{} \\
\Xhline{2\arrayrulewidth}
77.9   & 78.9   & 78.4 & 78.6  & 78.6    & 79.0 & \textbf{79.8} \\ 
\Xhline{2\arrayrulewidth}
\hline
\end{tabular}
\end{center}
\caption{\textbf{Few-shot classification accuracy on mini-ImageNet.}}
\label{table:few-shot}
\end{table}

To show OOD generalization beyond adversarial attacks, we compare SAGE to other data augmentation techniques for few-shot classification on mini-ImageNet, where the goal is to learn a representation that generalizes to novel categories.
We follow the setup from previous work~\citep{dvornik2019diversity}, using a single ResNet12 with the prototype classifier.
As shown in Table~\ref{table:few-shot}, \Ours{} outperforms other augmentation techniques, including Mixup (the strongest model on adversarial perturbations). 
This shows that \Ours{} is useful in OOD scenarios beyond Gaussian and adversarial perturbations.

\subsection{Runtime Analysis}\label{sec:runtime}
\begin{figure}[t]
\captionsetup[subfigure]{labelformat=simple, labelsep=period}
\centering 
\subfloat[Robustness Comparison]{\label{fig:cifar100_robustness_comparison}\includegraphics[width=.5\linewidth]{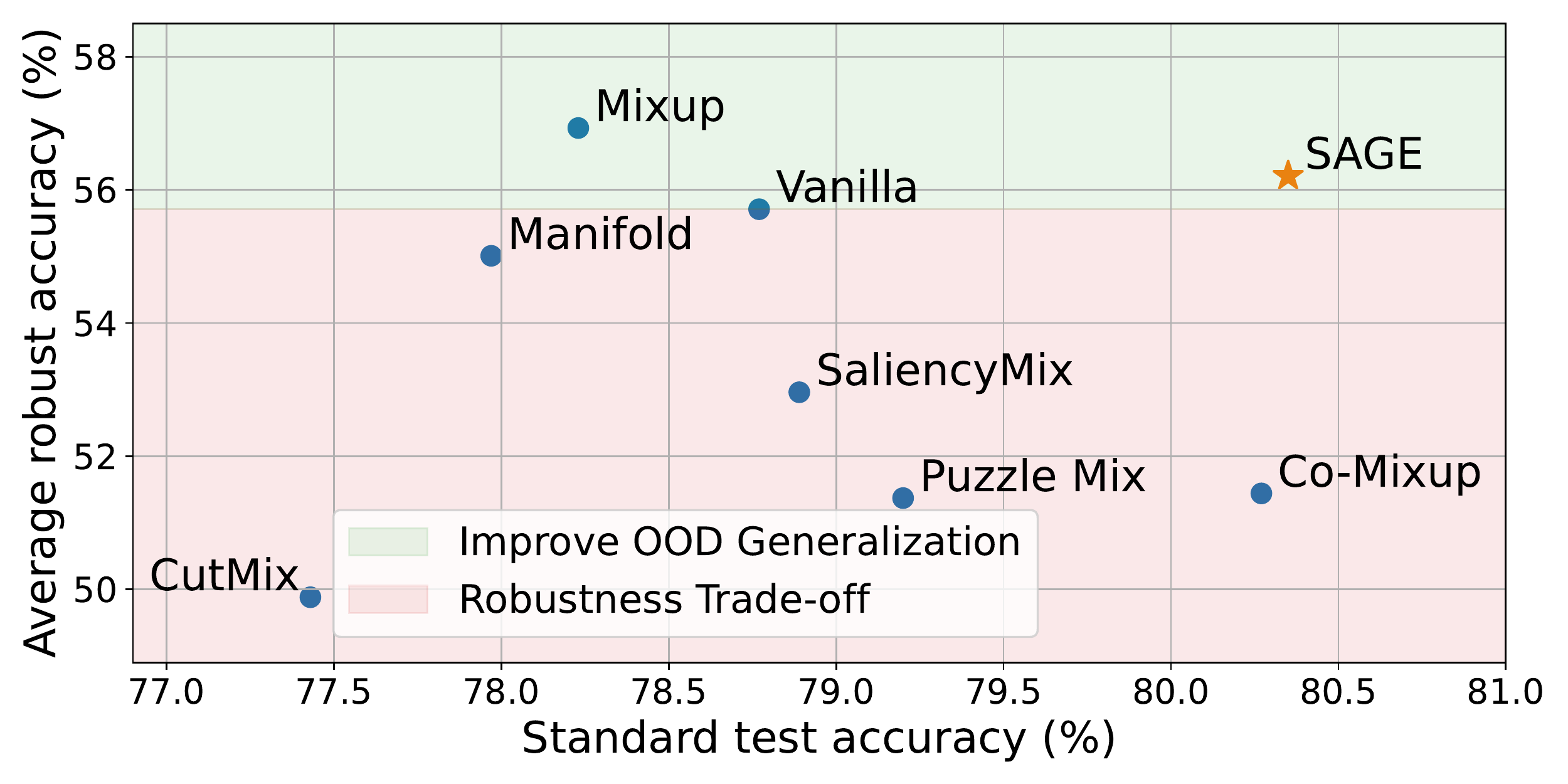}}
\hfill
\subfloat[Runtime Comparison]{\label{fig:cifar100_runtime_comparison}\includegraphics[width=.5\linewidth]{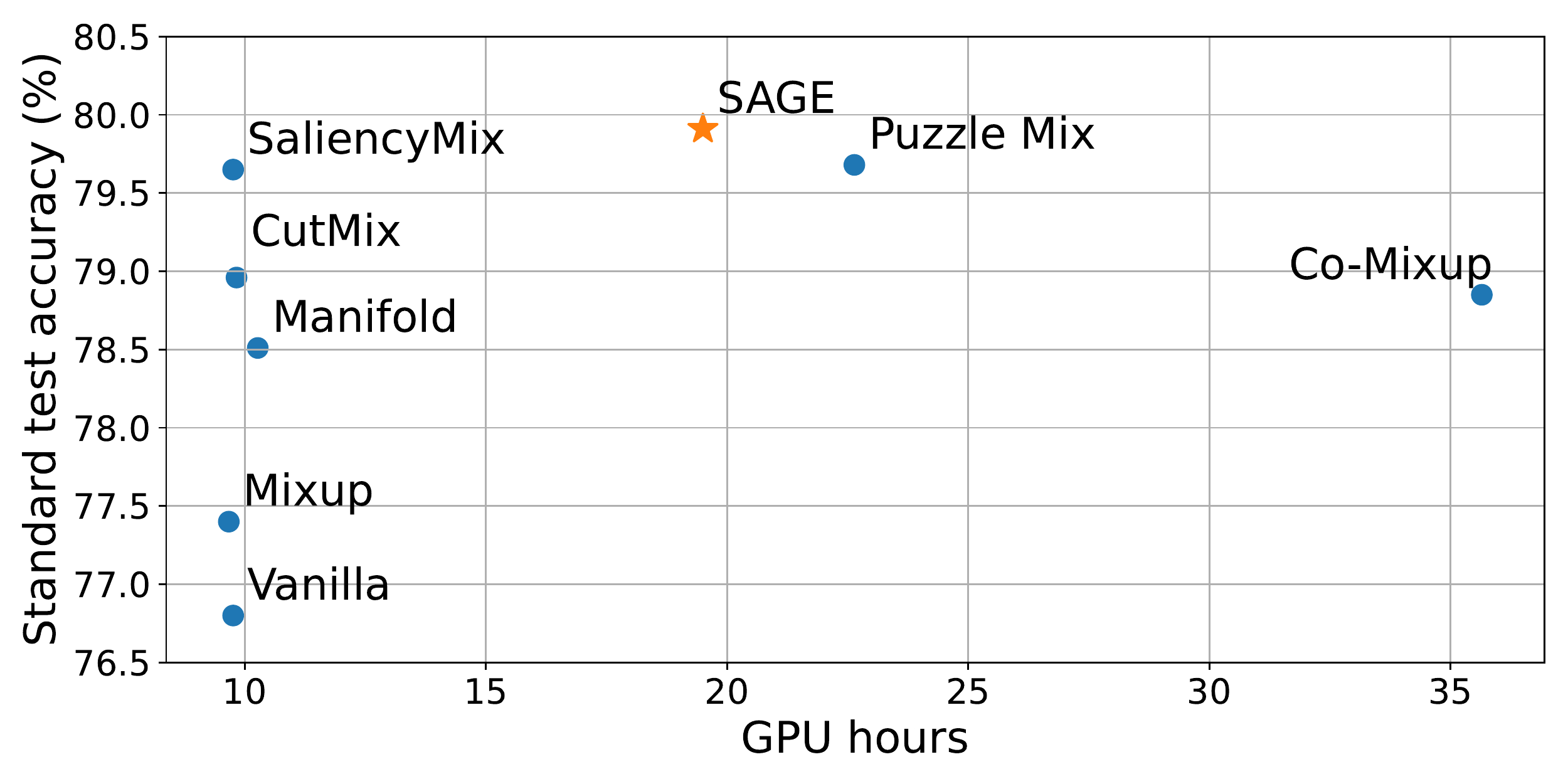}}
\caption{
\textbf{Robustness and efficiency analysis of \Ours{}.} 
(a) Robustness versus standard accuracy in OOD generalization. The methods in the green area (i.e., Mixup and \Ours{}) improve both accuracy and robustness relative to vanilla augmentation, while the others in red (i.e., CutMix, Co-Mixup and Puzzle Mix) improve standard test accuracy at the cost of decreased robustness.
(b) Runtime comparison of \Ours{} and other baselines. We estimate computation cost with a single NVIDIA Tesla T4. For SAGE, there is no noticeable overhead besides the additional forward and backward pass to compute the saliency map which approximately doubles the time of Vanilla training.
}
\label{fig:cifar100_runtime_robustness_comparison}
\end{figure}

In this section, we compare the training time of different data augmentation methods running on a single NVIDIA Tesla T4.
Figure~\ref{fig:cifar100_runtime_comparison}
plots each method's average training time (GPU hours) versus accuracy.
Notably, the techniques not using saliency (i.e., Mixup, Manifold and CutMix) are as fast as Vanilla, since the data augmentation is performed during data loading, which does not affect the overall training time.
SaliencyMix stands apart from the other saliency-based augmentation techniques. This follows because it utilizes an external trained saliency detector based on a shallow pre-deep learning method~\citep{montabone2010human}, that is fast but considerably less capable than the deep saliency methods~\cite{simonyan2013deep} used for the other augmentation techniques.
Consequently, SaliencyMix introduces minimal overhead; however, its improvement on classification accuracy is  limited. 
Other saliency-based methods (i.e., PuzzleMix, Co-Mixup and \Ours{}) are more accurate, yet also significantly slower.
Among them, \Ours{} is the fastest and also the most accurate on CIFAR-100.
Based on these observations, we argue that \Ours{} represents a good trade-off between accuracy and efficiency overall, and is clearly the best choice among the saliency-based methods.

\subsection{Ablation Studies}\label{sec:ablation}
In this section, we further analyze our data augmentation strategy by ablating different design choices in the pipeline.
For all the experiments, we use the same setup described in Sec.~\ref{sec:results} with ResNet18 on CIFAR-10 and CIFAR-100.  Please see the supplement for additional ablations.
%
%

\noindent\textbf{Saliency-guided Mixup and optimal rearrangements.}
The two components that make \Ours{} novel are the Saliency-guided Mixup (Sec.~\ref{sec:mixup}) and the Optimal Rearrangements (Sec.~\ref{sec:mix_with_shift}).
Here, we evaluate \Ours{} with some of the components removed or replaced by an existing technique.
In particular, we evaluate i) \Ours{} w/o OR (i.e., without optimal rearrangements) that always performs Saliency-guided Mixup on non-shifted images and ii) \Ours{} w/o SM (i.e., without Saliency-guided Mixup) for mixing images together that simply replaces one image region with the other image instead of performing smooth saliency-based mixing.
Examples of \Ours{} w/o SM and \Ours{} w/o OR are included in the supplement. 
As shown in Table~\ref{table:ablation_combined}, each of the components is important for the final performance and thus justifies their use.

\begin{table}[t]
\small
\centerline{
    \begin{minipage}{.5\linewidth}
       \hfill
       \begin{tabular}{lllll}
        Model                       & CIFAR-10 & CIFAR-100 \\
        \Xhline{2\arrayrulewidth}
        Vanila                      & 95.07             & 76.8          \\
        \Ours{} w/o SM              & 96.53             & 78.89             \\
        \Ours{} w/o OR              & 96.48             & 78.68         \\
        \Ours{}                     & \textbf{96.95}    & \textbf{79.91} \\
        \Xhline{2\arrayrulewidth}
        \end{tabular}
    \end{minipage}
    \hspace{4mm}
    \begin{minipage}{.5\linewidth}
       \begin{tabular}{ lllllll } 
                Search Space  & CIFAR-10 & CIFAR-100 \\
                \Xhline{2\arrayrulewidth}
                1\%      & \textbf{96.95}   & \textbf{79.91}       \\
                10\%     & 96.67            & 79.47         \\
                50\%     & 96.58            & 79.40         \\
                100\%    & 96.69            & 79.45         \\
                \Xhline{2\arrayrulewidth}
                \end{tabular}
        \hfill
    \end{minipage}}
\vspace{2mm}
\vspace{-1mm}
\caption{
\textbf{Ablation studies of \Ours{}.}
(left) Dissecting the benefit from saliency-guided mixing and optimal rearrangements.
Here, \Ours{} w/o SM (without Saliency-guided Mixup), and \Ours{} w/o OR (without optimal rearrangements).
(right) \Ours{}'s accuracy depending on the explored rearrangements. The first column indicates the size of the random portion of rearrangements used for data augmentation.
\vspace{-10pt}
}
\label{table:ablation_combined}
\end{table}

\noindent\textbf{Optimal rearrangements search space.}
As described in Sec.~\ref{sec:efficiency},
to select a rearrangement, we evaluate a set of locations, and proceed with the one that maximizes saliency. 
To speed up the search, we only explore a random subset of all rearrangements, 1\% in all previous experiments, which suggest that our data augmentation may be sub-optimal.
Table~\ref{table:ablation_combined} shows the model's performance, depending on the portion of all rearrangements we consider for DA.
Surprisingly, using only 1\% of the rearrangements works best.
While seemingly counterintuitive, we hypothesize the sub-optimal rearrangements act as additional training regularization and introduce more diversity in the augmented data.
\section{Conclusion}
\label{sec:conclusion}
We 
proposed \Ours{} -- a new data augmentation approach that integrates visual saliency to produce highly informative training samples.
Compared to existing methods, \Ours{} leads to better test accuracy, and generates more realistic training samples.
Moreover, \Ours{} is the only saliency-based augmentation technique that improves model robustness and OOD performance, while incurring minimal computational overhead.
In principle, \Ours{} is not limited to image classification and can be easily extended to other visual tasks. 
We believe that \Ours{} delivers a unique combination of accuracy, robustness and efficiency, and can become the new plug-and-play data augmentation for a wide range of vision tasks.

\newpage 
\bibliography{reference}
\bibliographystyle{plainnat}

\ifSupp
    \newpage
    \appendix
    \section{Summary of the Supplementary Material}
The supplementary material is organized as follows. 
In Sec.~\ref{sec:appendix:hyperparams}, we describe the exact optimization schedule and the hyperparameters used to train with SAGE and other baseline DA frameworks.
In Sec.~\ref{sec:appendix:robustness} and Sec.~\ref{sec:appendix:runtime}, we provide detailed results to bolster our claim on SAGE's improvement on OOD generalization (Sec. 4.2) and its low computation overhead (Sec. 4.3).
Pseudocode to augment data with SAGE is included in Sec.~\ref{sec:appendix:algorithm}.
In Sec.~\ref{sec:appendix:sage-sm-or}, we show examples of augmentations using SAGE w/o SM and SAGE w/o OR (Sec. 4.4).
Furthermore, we provide additional ablation studies to verify the design choices of SAGE in Sec.~\ref{sec:appendix:additional_ablations}.
\ifIncludeImageNet
    Finally, visualizations of SAGE augmented CIFAR-10 and ImageNet results are included in Sec.~\ref{sec:appendix:final_visualization}.
\fi

\section{Optimization schedule and hyper-parameters} \label{sec:appendix:hyperparams}
\textbf{Optimization schedule:} Following previous work~\cite{kim2020puzzle,kim2020co}, all models are trained using stochastic gradient descent (SGD) for 300 epochs with an initial learning rate of $0.2$. The learning rate decreases by a factor of 0.1 at epoch 100 and 200. We use a momentum of 0.9 and a weight decay of 0.0001. 
The above optimization schedule is used to train both CIFAR-10 and CIFAR-100 for all models, except for Co-Mixup~\cite{kim2020co} on CIFAR-10.
We notice that training with Co-Mixup on CIFAR-10 with an initial learning rate of 0.2 results in divergence at the beginning of the training. We find training becomes stable with an initial learning rate of 0.12.

\noindent\textbf{Training with baseline DA:} We follow the hyperparameter settings used in previous work \cite{kim2020puzzle, kim2020co}. To train with Mixup~\cite{zhang2018mixup}, CutMix~\cite{yun2019cutmix}, Puzzle Mix~\cite{kim2020puzzle} and Co-Mixup~\cite{kim2020co}, we use $\lambda \sim \text{Beta}(\alpha, \alpha)$ with $\alpha = 1.0$, and use $\alpha = 2.0$ for Manifold Mixup~\cite{verma2019manifold}.
For SaliencyMix~\footnote{\url{https://github.com/afm-shahab-uddin/SaliencyMix}}, Puzzle Mix\footnote{\url{https://github.com/snu-mllab/PuzzleMix}} and Co-Mixup\footnote{\url{https://github.com/snu-mllab/Co-Mixup}}, we use the parameter settings described in author's public repository: $(\beta, \mathbb{P}_{\text{mix}}) = (1.0, 0.5)$, $(\beta, \gamma, \eta, \eps) = (1.2, 0.5, 0.2, 0.8)$ and $(\beta, \gamma, \eta, \tau, \omega) = (0.32, 1.0, 0.05,0.83, 0.001)$.

\noindent\textbf{Training with SAGE:} For all models and datasets, we use $1\%$ of all possible rearrangements (Sec. 3.3) and a smoothing parameter of $\sigma^2=1.0$ (Sec. 3.2). 
Here we use $u$ to denote the truncation factor (Sec. \ref{sec:appendix:additional_ablations}) and use $\eta$ to denote the gradient update ratio (Sec. 3.4).
On CIFAR-10 with ResNet18, we $(u, \eta) = (0.6, 0.7)$. On CIFAR-100 with ResNet18, we $(u, \eta) = (0.5, 0.7)$. On CIFAR-100 with WRN16, we $(u, \eta) = (0.6, 0.7)$. On CIFAR-100 with ResNext29, we $(u, \eta) = (0.7, 0.5)$.

\section{Robustness Evaluation on CIFAR-10 and CIFAR-100} \label{sec:appendix:robustness}
We evaluate the robustness of models trained with various baseline DA methods. In particular, we measure the classification accuracy of models on test data perturbed using Gaussian noise ($\sigma^2=0.01$) and adversarial attacks. To craft adversarial perturbations, we use $\eps = \frac{8}{255}$ for $\ell_{\infty}$ bounded FGSM~\cite{goodfellow2014explaining} and $\eps= 0.5$ for $\ell_{2}$ bounded FGM~\cite{goodfellow2014explaining}. Results are based on models trained with ResNet18.
In Figure~\ref{fig:robustness_tradeoff}, we notice that models trained with \Ours{} achieve improved classification accuracy on both clean and noise-perturbed test data. 
However, method such as SaliencyMix, Puzzle Mix, Co-Mixup and CutMix improves generalization performance on the test data at the cost of decreased robustness.

\begin{table}[htbp]
\footnotesize
\begin{center}
\setlength{\tabcolsep}{5pt} 
\renewcommand{\arraystretch}{1.3} 
\begin{tabular}{ lllllllll} 
Perturbations                 & Vanilla     & Mixup     & CutMix   & Manifold   & SaliencyMix         & Puzzle Mix   & Co-Mixup      & \Ours{} \\ \Xhline{2\arrayrulewidth}
Rank           &4 &3  &8  &1  &6  &5 &7  & 2 \\
FGSM ($\ell_{\infty}$)          & 79.96       & 80.93     & 79.57    & \textbf{85.79}  & 80.62    & 81.96     & 78.78   & \underline{83.75} \\ 
FGM ($\ell_{2}$)              & 89.67       & 89.22     & 87.81    & \textbf{90.86}    & 88.86           & 89.64   & 88.11    & \underline{90.64} \\ 
Gaussian               & 89.88       & \textbf{92.56}     & 77.2     & \underline{92.21}  & 85.99    & 87.60     & 85.25   & 91.67 \\ \Xhline{2\arrayrulewidth}
\hline
\end{tabular}
\end{center}
\caption{Classification accuracy on noise perturbed CIFAR-10 test data.}
\label{table:robustness}
\end{table}

\begin{table}[htbp]
\footnotesize
\begin{center}
\setlength{\tabcolsep}{5pt} 
\renewcommand{\arraystretch}{1.3} 
\begin{tabular}{ lllllllll} 
Perturbations                      & Vanilla     & Mixup   & CutMix   & Manifold & SaliencyMix  & Puzzle Mix    & Co-Mixup       & \Ours{} \\ \Xhline{2\arrayrulewidth}
Rank &3 &1 &8 &4 &5 &7 &6 &2 \\
FGSM ($\ell_{\infty}$)         & 49.24       & \textbf{50.51}     & 44.2    & 48.89   & 46.52   & 44.58     & 44.32   & \underline{50.18} \\ 
FGM ($\ell_{2}$)             & 62.19       & \textbf{63.36}     & 55.57    & 61.14  & 59.4    & 58.16     & 58.56    & \underline{62.23} \\ 
Gaussian              & 52.68       & \textbf{60.76}     & 28.06     & \underline{55.47}   & 38.21   & 43.96     & 34.46   & 47.68 \\ \Xhline{2\arrayrulewidth}
\hline
\end{tabular}
\end{center}
\caption{Classification accuracy on noise perturbed CIFAR-100 test data. }
\label{table:robustness_cifar100}
\end{table}

\begin{figure}[htbp]
\captionsetup[subfigure]{labelformat=empty}
\centering 
\subfloat[CIFAR-10]{\includegraphics[width=0.5\linewidth]{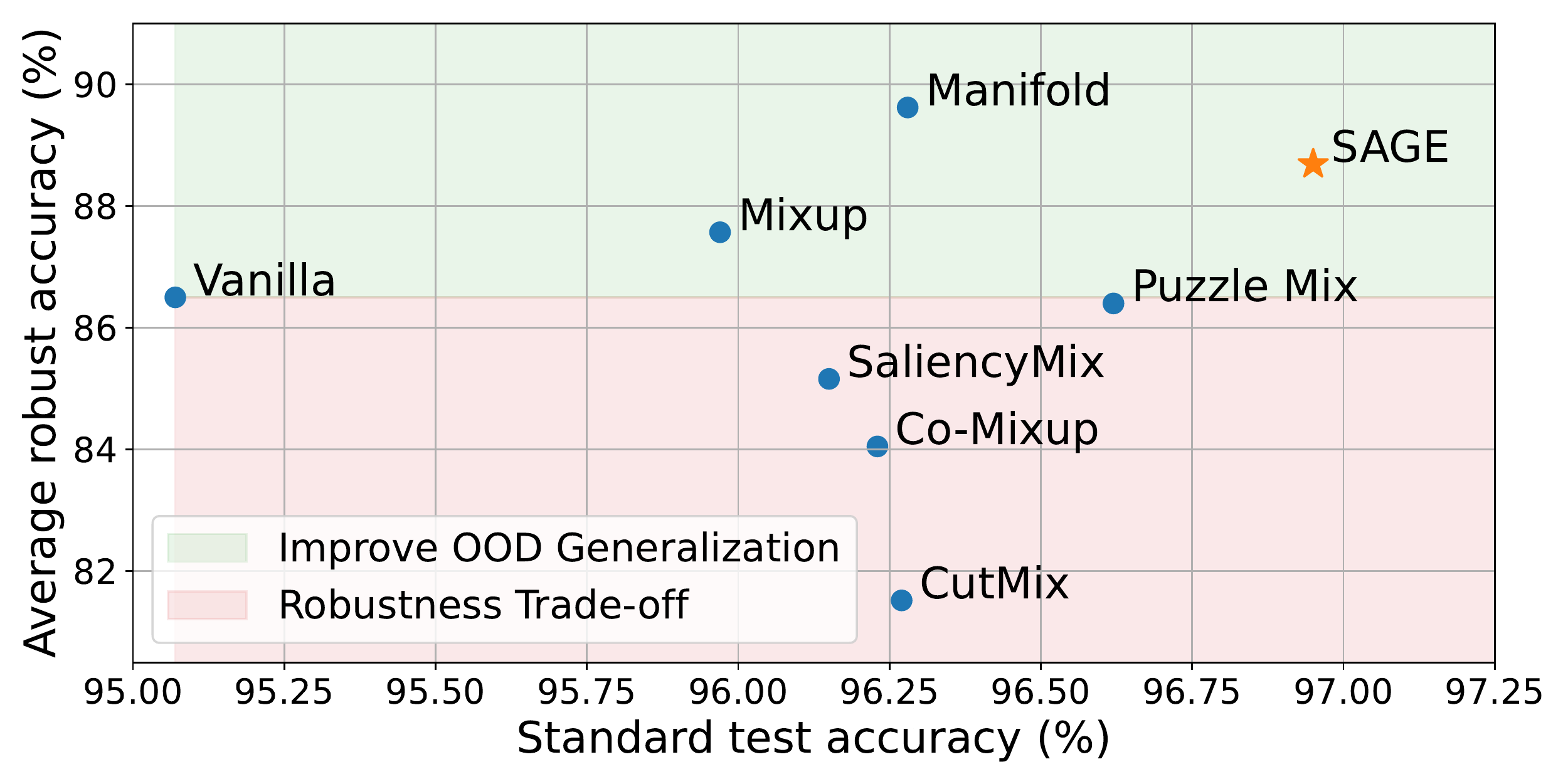}}
\hfill
\subfloat[CIFAR-100]{\includegraphics[width=0.5\linewidth]{figure/robustness-cifar100_tall.pdf}}
\caption{Visualization of the standard generalization performance vs. generalization in the OOD setting. We notice that SaliencyMix, CutMix, Co-Mixup and Puzzle Mix improves standard test accuracy over vanilla but at a cost of decreased robustness.
}
\label{fig:robustness_tradeoff}
\end{figure}




\section{Runtime Comparison}\label{sec:appendix:runtime}
To estimate the computation cost of various baseline DA methods, we measure the total GPU hours required to train CIFAR-10 and CIFAR-100 using a single NVIDIA Tesla T4. Notice training with \Ours{} approximately doubles the time of vanilla training due to the computation of the saliency map; however, unlike Puzzle Mix and Co-Mixup, there is no additional overhead in finding the optimal rearrangements to maximize the total saliency.
SaliencyMix stands apart from the other saliency-based augmentation techniques. This follows because it utilizes an external trained saliency detector based on a shallow pre-deep learning method~\citep{montabone2010human}, that is fast but considerably less capable than the deep saliency methods~\cite{simonyan2013deep} used for the other augmentation techniques.
Consequently, SaliencyMix introduces minimal overhead; however, its improvement on classification accuracy is  limited. 

\begin{table}[htbp]
\footnotesize
\begin{center}
\setlength{\tabcolsep}{2pt} 
\renewcommand{\arraystretch}{1.3} 
\begin{tabular}{ llllll|llll } 
Dataset & Model                        & Vanilla     & Mixup   & CutMix   & Manifold  & SaliencyMix & Puzzle Mix   & Co-Mixup       & \Ours{} \\ \Xhline{2\arrayrulewidth}
CIFAR10 & PreActResNet18             &3.35      & 3.29 & 3.38 & 3.44 & 3.45 & 8.9 & 25.29    &\textbf{6.83} \\ 
CIFAR100 & ResNext29     &9.76      & 9.67 & 9.83 & 10.27 & 10.18 & 22.64 & 35.65    &\textbf{19.5} \\ \Xhline{2\arrayrulewidth}
\end{tabular}
\end{center}
\caption{GPU hours comparison of \Ours{} and other baselines.}
\label{table:runtime_comparison}
\end{table}
\vspace{-20pt}
\begin{figure}[htbp]
\captionsetup[subfigure]{labelformat=empty}
\centering 
\subfloat[CIFAR-10]{\includegraphics[width=0.5\linewidth]{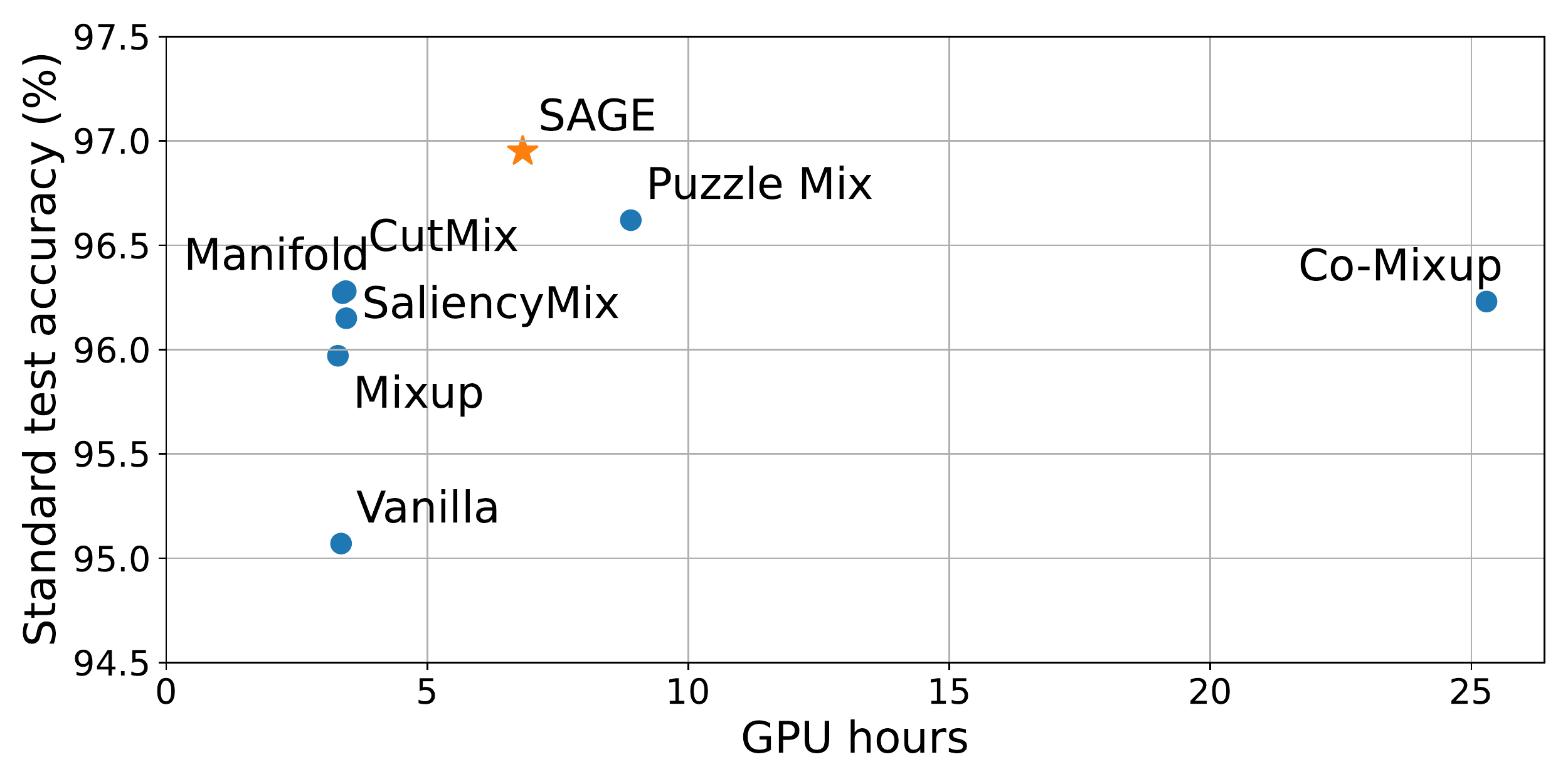}}
\hfill
\subfloat[CIFAR-100]{\includegraphics[width=0.5\linewidth]{figure/runtime-cifar100_tall.pdf}}
\caption{Compared to other saliency-guided methods, \Ours{} achieves better standard test accuracy on both datasets with low computation overhead.
}
\label{fig:runtime_comparison}
\end{figure}
\vspace{-20pt}
\section{Full SAGE Algorithm}\label{sec:appendix:algorithm}

Algorithm~\ref{algorithm} shows the exact procedure of \Ours{}. We discuss saliency-guided mixing with optimal rearrangement (Ln 3) in Sec. 3.2, and the rest of the algorithm is covered in Sec. 3.1. 

\begin{algorithm}[htbp]
  \SetKwData{Left}{left}\SetKwData{This}{this}\SetKwData{Up}{up}
    \SetKwInOut{Input}{Input}\SetKwInOut{Output}{Output}
    \Input{Pairs of training samples: $(x_0,y_0)$ and $(x_1,y_1)$, a classifier $f_\theta(\cdot)$, a loss function $\ell$, a randomly sampled mix ratio $\lambda$, a Gaussian smoothing parameter $\sigma^2$ and $\mathcal{O}$ is the space of all possible image translations}
    \Output{A new data-label pair: $(x',y')$}
    \BlankLine
    
    \SetAlgoNoLine
    
    $s_0 = \abs{ \nabla_{x} \ell(f_{\theta}(x_0), y_0) }_{l_{2,3}}, s_1 = \abs{ \nabla_{x} \ell(f_{\theta}(x_1), y_1) }_{l_{2,3}}$\\
    $\tilde{s}_0 = \lambda*\text{Smoothing}(s_0, \sigma^2),\ \tilde{s}_1 = (1-\lambda)*\text{Smoothing}(s_1, \sigma^2)$\\
    $ \tau^*  = \underset{\tau \in \mathcal{O}}{\argmax}\; \nu(\tau)$, where $\nu(\tau)$ is defined in Eq. 4 \\
    $ M^{\tau^*} = \frac{\tilde{s}_0}{\tilde{s}_0 + \mathcal{T}(\tilde{s}_1; \tau^*)+\zeta}$\\
    $ \gamma = \frac{1}{d^2}\sum^{d}_{i,j=1}M^{\tau^*}_{ij}$\\
    $ x' = M^{\tau^*}\odot x_0 + (1-M^{\tau^*})\odot \mathcal{T}(x_1; \tau^*)$\\
    $ y' = \gamma \cdot y_0 + (1-\gamma) \cdot y_1$\\
            
    \caption{Data Augmentation based on SMART Mixup}
\label{algorithm}
\end{algorithm}

\section{Examples of Augmentation Results with SAGE w/o SM and SAGE w/o OR}\label{sec:appendix:sage-sm-or}
In Sec. 4.4, we verified the effectiveness of our data augmentation strategy by ablating i) \Ours{} w/o OR (i.e., without optimal rearrangements) that always performs Saliency-guided Mixup on non-shifted images and ii) \Ours{} w/o SM (i.e., without Saliency-guided Mixup). Examples of the augmentation results are shown in Figure~\ref{fig:sage-sm-and-sage-or}.

\begin{figure}[htbp]
\captionsetup[subfigure]{labelformat=empty}
\centering 
\subfloat[]{\includegraphics[width=0.23\linewidth]{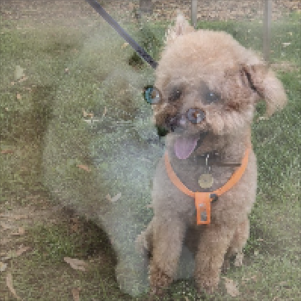}}
\hfill
\subfloat[]{\includegraphics[width=0.23\linewidth]{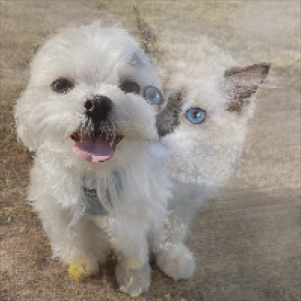}}
\hfill
\subfloat[]{\includegraphics[width=0.23\linewidth]{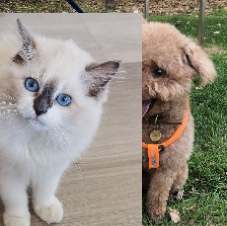}}
\hfill
\subfloat[]{\includegraphics[width=0.23\linewidth]{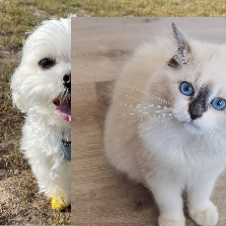}}
\\
\vspace{-5pt}
SAGE w/o OR \hspace{140pt} SAGE w/o SM
\caption{Augmentation results with SAGE w/o OR and SAGE w/o SM}
    \label{fig:sage-sm-and-sage-or}
\end{figure}
\vspace{-10pt}
\section{Additional Ablations} \label{sec:appendix:additional_ablations}
We include two additional ablation studies in this section: i) reusing parameter gradients from un-augmented samples (Sec. 3.4) and ii) randomly rescaling of total saliency.
For each experiment, we use the best result as the control group (bold numbers), then we repeat the runs with modified task-related parameters.

\begin{table}[h]
\small
\centerline{
    \begin{minipage}{.5\linewidth}
        \hfill
       \begin{tabular}{ lllllll } 
                $\eta$  & CIFAR-10 & CIFAR-100 \\
                \Xhline{2\arrayrulewidth}
                0.5     & 96.65            & 79.36       \\
                0.7     & \textbf{96.95}            & \textbf{79.91}         \\
                1.0     & 96.58            & 79.24         \\
                \Xhline{2\arrayrulewidth}
                \end{tabular}
        
    \end{minipage}
    \hspace{4mm}
    \begin{minipage}{.5\linewidth}
       \begin{tabular}{lllll}
        $u$                       & CIFAR-10 & CIFAR-100 \\
        \Xhline{2\arrayrulewidth}
        0.5                      & 96.75             & \textbf{79.91}          \\
        0.6                      & \textbf{96.95}             & 79.7             \\
        1.0                      & 96.6             & 79.29         \\
        \Xhline{2\arrayrulewidth}
        \end{tabular}
        \hfill
    \end{minipage}}
\vspace{2mm}
\vspace{-1mm}
\caption{
\textbf{Additional ablation studies of \Ours{}.}
(left) Test accuracy of models trained with combined parameter gradients from un-augmented and SAGE-augmented samples.
(right) Test accuracy of models trained with truncated total saliency.
\vspace{-10pt}
}
\label{table:additional_ablation}
\end{table}

\textbf{Reusing the parameter gradients:} 
In Sec. 3.4, we discuss performing gradient descent update by combining parameter gradients computed on un-augmented and SAGE-augmented samples. 
In particular, let $g_s$ and $g_a$ represent the gradients computed using un-augmented and augmented images, respectively. The final model update is based on $g = \eta \cdot g_s + (1-\eta) \cdot g_a$, where $\eta \in [0, 1]$.
%
In Table~\ref{table:additional_ablation}, we observe reusing the parameter gradients computed on un-augmented samples ($\eta \neq 1$) significantly increases accuracy on the test data.

\textbf{Random rescaling of total saliency:} A random mixing ratio in prior work~\cite{zhang2018mixup, yun2019cutmix, devries2017improved, kim2020co, kim2020puzzle} can be seen as a way to increase diversity of the augmentation results.
Similarly, we randomly rescale the total saliency of smoothed $s_0$ and $s_1$ using $\lambda \sim \mathcal{U}(0, 1)$ and $1-\lambda$ respective.
In practice, we observe the diversity in the augmented images greatly decreases when $\lambda > 0.6$, since $x_0$ and $\tilde{s}_0$ dominate when computing the total saliency. 
Therefore, when the offset images are rescaled to having a small total saliency, it is often better to just exclude it in the augmented results.
\ifIncludeImageNet
    We include a detailed comparison of augmented images using $\lambda$ linearly increasing from 0 to 1 in Figure~\ref{fig:cifar10_results} and Figure~\ref{fig:imagenet_results}.
\fi
As such, we propose a simple heuristic to truncate the random rescaling factor: $\lambda \sim \mathcal{U}(0, u)$, where $u \in [0, 1]$. 
Results in Table~\ref{table:additional_ablation} shows with $u < 1.0$, the test accuracy on both datasets increase significantly.

\ifIncludeImageNet
    \newpage
    \section{Examples of SAGE Augmented Images} \label{sec:appendix:final_visualization}
    \begin{figure}[thbp]
        \begin{tabular}{l l}
        $\lambda=0$ & \includegraphics[height=\FigSuppHeight mm]{figure/truncation/cifar10_lambda-0.png}
        \vspace{-5pt}\\
        $\lambda=0.1$ & \includegraphics[height=\FigSuppHeight mm]{figure/truncation/cifar10_lambda-1.png}
        \vspace{-5pt}\\
        $\lambda=0.2$ & \includegraphics[height=\FigSuppHeight mm]{figure/truncation/cifar10_lambda-2.png}
        \vspace{-5pt}\\
        $\lambda=0.3$ & \includegraphics[height=\FigSuppHeight mm]{figure/truncation/cifar10_lambda-3.png}
        \vspace{-5pt}\\
        $\lambda=0.4$ & \includegraphics[height=\FigSuppHeight mm]{figure/truncation/cifar10_lambda-4.png}
        \vspace{-5pt}\\
        $\lambda=0.5$ & \includegraphics[height=\FigSuppHeight mm]{figure/truncation/cifar10_lambda-5.png}
        \vspace{-5pt}\\
        $\lambda=0.6$ & \includegraphics[height=\FigSuppHeight mm]{figure/truncation/cifar10_lambda-6.png}
        \vspace{-5pt}\\
        $\lambda=0.7$ & \includegraphics[height=\FigSuppHeight mm]{figure/truncation/cifar10_lambda-7.png}
        \vspace{-5pt}\\
        $\lambda=0.8$ & \includegraphics[height=\FigSuppHeight mm]{figure/truncation/cifar10_lambda-8.png}
        \vspace{-5pt}\\
        $\lambda=0.9$ & \includegraphics[height=\FigSuppHeight mm]{figure/truncation/cifar10_lambda-9.png}
        \vspace{-5pt}\\
        $\lambda=1.0$ & \includegraphics[height=\FigSuppHeight mm]{figure/truncation/cifar10_lambda-9.png}
        \end{tabular}
        \caption{SAGE augmented images with linearly increasing $\lambda$ on CIFAR-10.}
        \label{fig:cifar10_results}
    \end{figure}
    \begin{figure}[thbp]
        \begin{tabular}{l l}
        $\lambda=0$ & \includegraphics[height=\FigSuppHeight mm]{figure/truncation/tiny_lambda-0.png}
        \vspace{-5pt}\\
        $\lambda=0.1$ & \includegraphics[height=\FigSuppHeight mm]{figure/truncation/tiny_lambda-1.png}
        \vspace{-5pt}\\
        $\lambda=0.2$ & \includegraphics[height=\FigSuppHeight mm]{figure/truncation/tiny_lambda-2.png}
        \vspace{-5pt}\\
        $\lambda=0.3$ & \includegraphics[height=\FigSuppHeight mm]{figure/truncation/tiny_lambda-3.png}
        \vspace{-5pt}\\
        $\lambda=0.4$ & \includegraphics[height=\FigSuppHeight mm]{figure/truncation/tiny_lambda-4.png}
        \vspace{-5pt}\\
        $\lambda=0.5$ & \includegraphics[height=\FigSuppHeight mm]{figure/truncation/tiny_lambda-5.png}
        \vspace{-5pt}\\
        $\lambda=0.6$ & \includegraphics[height=\FigSuppHeight mm]{figure/truncation/tiny_lambda-6.png}
        \vspace{-5pt}\\
        $\lambda=0.7$ & \includegraphics[height=\FigSuppHeight mm]{figure/truncation/tiny_lambda-7.png}
        \vspace{-5pt}\\
        $\lambda=0.8$ & \includegraphics[height=\FigSuppHeight mm]{figure/truncation/tiny_lambda-8.png}
        \vspace{-5pt}\\
        $\lambda=0.9$ & \includegraphics[height=\FigSuppHeight mm]{figure/truncation/tiny_lambda-9.png}
        \vspace{-5pt}\\
        $\lambda=1.0$ & \includegraphics[height=\FigSuppHeight mm]{figure/truncation/tiny_lambda-9.png}
        \end{tabular}
        \caption{SAGE augmented images with linearly increasing $\lambda$ on ImageNet.}
        \label{fig:imagenet_results}
    \end{figure}
\fi
\else
\fi

\end{document}